\documentclass[lettersize,journal]{IEEEtran}
\usepackage{amsmath,amsfonts}
\usepackage{algorithmic}
\usepackage{algorithm}
\usepackage{array}
\usepackage[caption=false,font=normalsize,labelfont=sf,textfont=sf]{subfig}
\usepackage{textcomp}
\usepackage{stfloats}
\usepackage{url}
\usepackage{verbatim}
\usepackage{graphicx}
\usepackage{cite}

\usepackage{multirow}
\usepackage{makecell}
\usepackage{hyperref}
\usepackage[hyphenbreaks]{breakurl}
\usepackage{booktabs} 
\usepackage{float}

\begin{document}

\title{MUSTER: A Multi-scale Transformer-based Decoder for Semantic Segmentation}

\author{Jing Xu\textsuperscript{\textsection}, Wentao Shi\textsuperscript{\textsection}, Pan Gao$^*$, Qizhu Li, and Zhengwei Wang
\thanks{This work was supported by  the Natural Science Foundation of China under Grant 62272227. \emph{(Corresponding author: Pan Gao.)}}
\thanks{Jing Xu, Wentao Shi and Pan Gao  are with the College
of Computer Science and Technology, Nanjing University of Aeronautics and Astronautics, Nanjing 211106, China.
}
\thanks{Qizhu Li is with TikTok Pte.~Ltd.,~Singapore.}
\thanks{Zhengwei Wang is with ByteDance, Shanghai 201103, China.}

\thanks{\textsuperscript{\textsection}Equal contribution.}

\thanks{© 2024 IEEE.  Personal use of this material is permitted.  Permission from IEEE must be obtained for all other uses, in any current or future media, including reprinting/republishing this material for advertising or promotional purposes, creating new collective works, for resale or redistribution to servers or lists, or reuse of any copyrighted component of this work in other works.}
}



\maketitle

\begin{abstract}
    In recent works on semantic segmentation, there has been a significant focus on designing and integrating transformer-based encoders. However, less attention has been given to transformer-based decoders. We emphasize that the decoder stage is equally vital as the encoder in achieving superior segmentation performance. It disentangles and refines high-level cues, enabling precise object boundary delineation at the pixel level. In this paper, we introduce a novel transformer-based decoder called MUSTER, which seamlessly integrates with hierarchical encoders and consistently delivers high-quality segmentation results, regardless of the encoder architecture. Furthermore, we present a variant of MUSTER that reduces FLOPS while maintaining performance. MUSTER incorporates carefully designed multi-head skip attention (MSKA) units and introduces innovative upsampling operations. The MSKA units enable the fusion of multi-scale features from the encoder and decoder, facilitating comprehensive information integration. The upsampling operation leverages encoder features to enhance object localization and surpasses traditional upsampling methods, improving mIoU (mean Intersection over Union) by 0.4\% to 3.2\%. 
    On the challenging ADE20K dataset, our best model achieves a single-scale mIoU of 50.23 and a multi-scale mIoU of 51.88, which is on-par with the current state-of-the-art model. Remarkably, we achieve this while significantly reducing the number of FLOPs by 61.3\%.
   Our source code and models are publicly available at: \href{https://github.com/shiwt03/MUSTER}{https://github.com/shiwt03/MUSTER}
\end{abstract}


\begin{figure}[htb]
    \centering
    \includegraphics[width=\columnwidth]{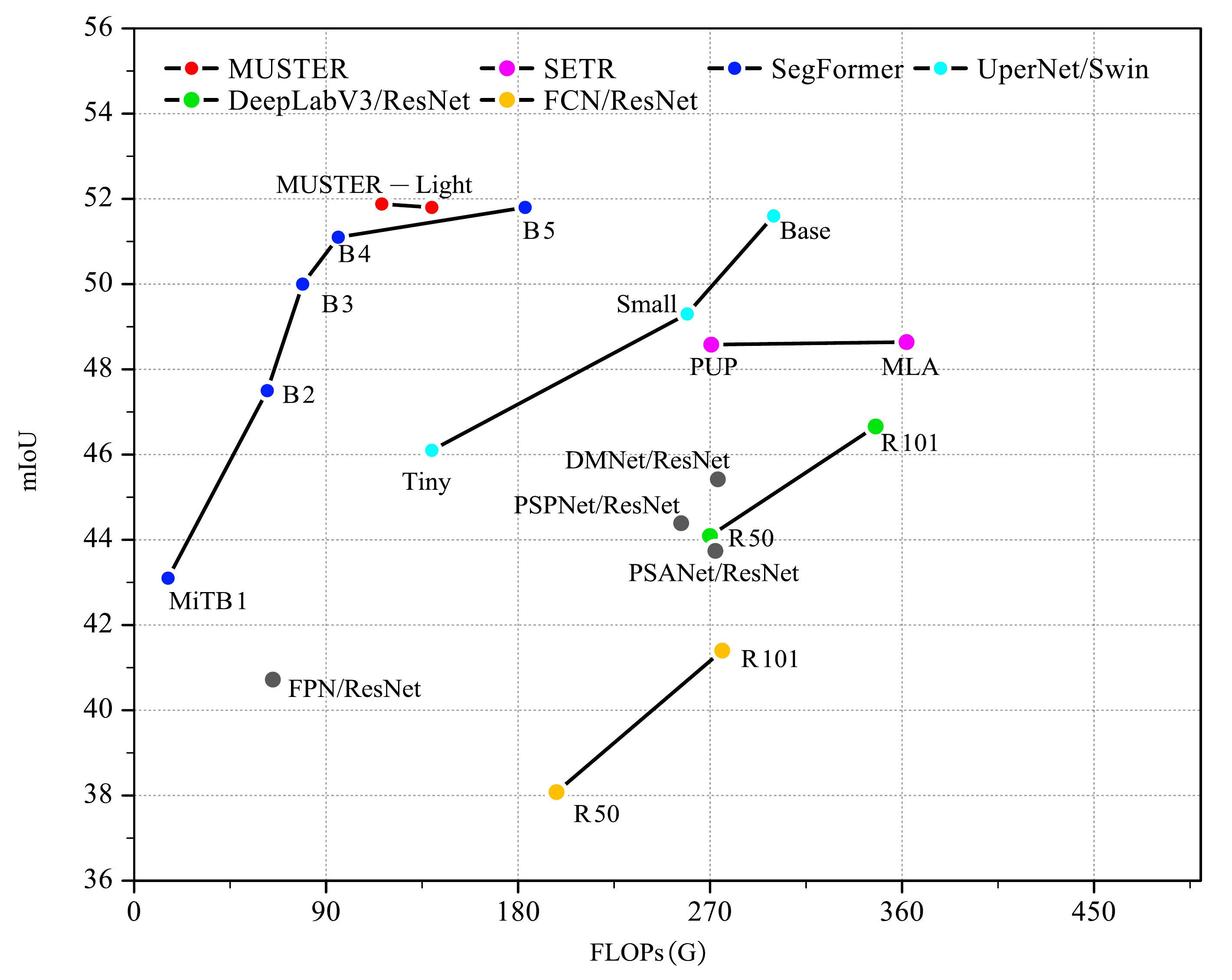}
    \caption{mIoU vs FLOPs on ADE20K dataset. Our MUSTER achieves 8.9\% higher mIoU than original Swin Transformer with same FLOPs, and cuts 53.5\% FLOPs while remaining almost the same mIoU on ADE20K.}
    \label{fig9}
\end{figure}

\begin{figure*}[t]
    \centering
    \includegraphics[width=\linewidth]{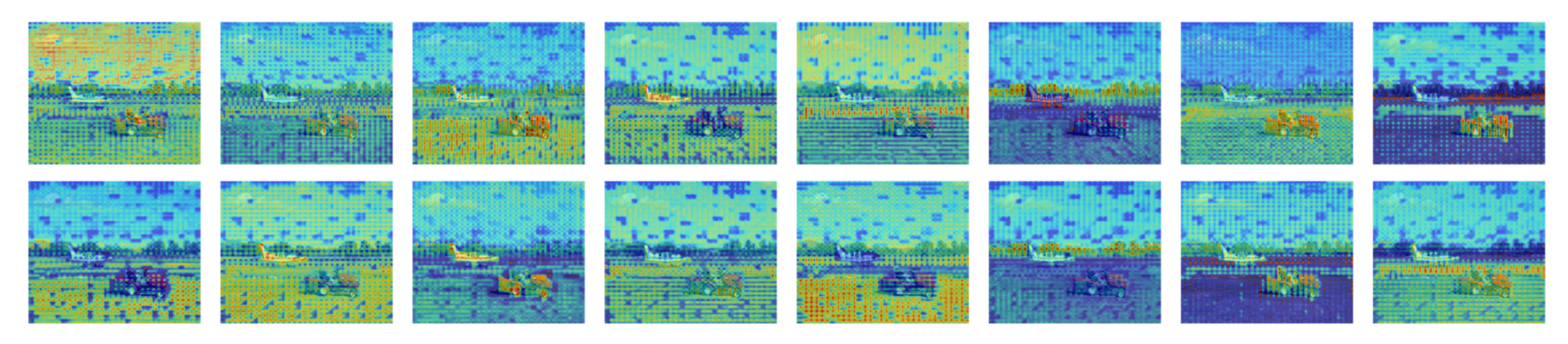}
    \caption{Feature maps of 16 heads of the second stage in MUSTER, where red and orange areas are the focus of a single head.}
    \label{fig9}
\end{figure*}

\section{Introduction}
\label{sec:intro}
Semantic segmentation is a task that predicts the category of each pixel in an image, rather than the image level prediction in image classification task. In semantic segmentation, the decoder is as important as the encoder, or even more so, because it is more challenging to upsample a sequence to a segmentation map than to obtain features from the sequence.

Compared to traditional convolutional neural networks (CNN) \cite{FCN}, it has been shown that transformers  perform better in computer vision tasks including semantic segmentation~\cite{ViT,Swin,segmenter,Trseg}. Nevertheless, existing works only pay attention to the application of the transformer or its variants as a backbone in a model in order to diffusely use it in all computer vision or even multimedia tasks.
We believe a well-designed decoder is critical to realizing the full potential of strong backbones, as the encoder serves as the sole link between the wealth of information extracted by the backbone, and the desired prediction. There are a number of well-designed segmentation decoders in CNN-based networks, such as FPN \cite{FPN}, FCN \cite{FCN}, and UperNet \cite{Upernet}. 
However, there are also some weaknesses that should be considered: CNN decoders are not as effective as transformer decoders in fully understanding the feature maps extracted by the transformer backbone. This is because CNN decoders rely on a limited receptive field to decode feature maps, which may not capture all relevant information.
Another weakness of CNN decoders is their lack of flexibility in adapting to different input sizes and aspect ratios. This can lead to reduced accuracy when processing images with varying scales or orientations.
In addition, CNN decoders typically require a large number of floating point operations (e.g., convolution and bilinear) to achieve good performance, which can make them computationally expensive and difficult to train on limited hardware resources.
After rethinking the applications of Swin Transformer~\cite{Swin} in semantic segmentation, in this paper, we design an innovative \textbf{MU}ti-\textbf{S}cale \textbf{T}ransformer-based  decod\textbf{ER}, dubbed MUSTER, in order to not only address the limited receptive field and high complexity issues of CNN decoders, but also propose a generic and performant method applicable to any other types of hierarchical backbones in the field of semantic segmentation. 

While a large number of existing works focus on the backbone design, few methods \cite{ANN,FCN} in semantic segmentation pay attention to designing the effective skip connection operation. However, due to the dense prediction characteristic of segmentation, it is crucial to obtain local features from a high resolution feature image. We recognize recent works (e.g., U-Net~\cite{UNetCN} and SegFormer~\cite{SegFormer}) alleviate this issue by utilizing simple concatenation. However, the concatenation is not able to fully capture local features, since directly concatenating will induce many ineffective and redundant information from encoder features. To tackle this problem, we propose a novel method using multi-head skip attention (MSKA) to more efficiently aggregate the effective feature information from the encoder.

When applying the transformer~\cite{Attention} from natural language processing to computer vision such as image classification \cite{ViT}, the original decoder of transformer is usually removed, and instead an MLP layer is added as the prediction head. This is because that, unlike the NLP task such as machine translation that needs to model the input and output sequences via the encoder and decoder, respectively, image recognition does not involve an output sequence, and can  output probability directly from the input sequence of image patches.
In this paper, considering the example of transformer decoder, we design a Swin Transformer-symmetric segmentation head to merge different feature information. We call our proposed attention mechanism \textbf{M}ultihead \textbf{SK}ip \textbf{A}ttention (MSKA), since it originates from implementing skip connection using multi-head self-attention. Feature maps of each head of MSKA are shown in Figure \ref{fig9}, which clearly indicates different localized areas done by the 16 heads of MSKA.

Our objective is to design a decoder that excels in efficiency while adhering to stringent constraints on computational resources. Consequently, we introduce Light-MUSTER, a refined variant of MUSTER that is meticulously optimized for maximum performance within a lightweight framework. To preserve the fine structure while reducing the complexity, we modify the attention method by removing the linear projection layers based on ViT-LSLA\cite{Vitlsla}. We also downsample the internal feature map and redesign the mask mechanism in the shift window multi-head skip attention to lower the flops in matrix multiplication.  Moreover, our proposed decoder can be widely and simply used with any pyramid/hierarchical backbones, even CNN-based models, e.g., ResNet\cite{resnet}. 

Compared to existing methods like UperNet\cite{Upernet}, SegFormer\cite{SegFormer}, SETR\cite{setr}, and FCN\cite{FCN}, our MUSTER method offers significant advantages. It integrates a global receptive field through a transformer-based decoder and leverages a Unet-shaped architecture for efficient multi-scale feature integration. The Multihead Skip Attention (MSKA) mechanism enhances feature fusion, improving segmentation accuracy while reducing redundancy.

The major contributions of this paper are summarized as follows:
\begin{itemize}
\item{Firstly, inspired by the symmetric encoder and decoder architecture in CNN, we propose a hierarchical  transformer-based decoder for semantic segmentation, dubbed MUSTER, which gradually increases the spatial resolution of the features based on attention mechanism. The MUSTER does the reverse operation as Swin Transformer, which enables the use of pretraining of Swin Transformer for fine-tuning, distinctly different from existing  transformer-based segmentation heads (e.g., SegFormer \cite{SegFormer} and PVT \cite{PyramidVT}) which need to pretrain the encoder.}
\item{To enable dense per-pixel category prediction, we propose a \textbf{M}ultihead \textbf{SK}ip \textbf{A}ttention (MSKA) block, which combines the upsampled output at the decoder and the same-scale feature at the encoder via self-attention mechanism. As will be demonstrated, this skip connection via self-attention exhibits better localization performance compared to the concatenation operation, which is adopted almost in all current  encoder-decoder symmetric structures, e.g., CNN-based U-Net \cite{UNetCN} and SegNet \cite{SegNetAD}, transformer-based TransUnet \cite{TransUNet,wang2022uctransnet,cao2021swin}.}
\item{We design a lightweight version of MUSTER to lower the model complexity and boost the inference efficiency. Instead of simply removing some blocks or downsampling the feature map in a crude way, we modify the attention mechanism and the entire decoder architecture.}
\end{itemize}

\section{Related Works}

\begin{figure*}[ht]
    \centering
    \includegraphics[width=\textwidth]{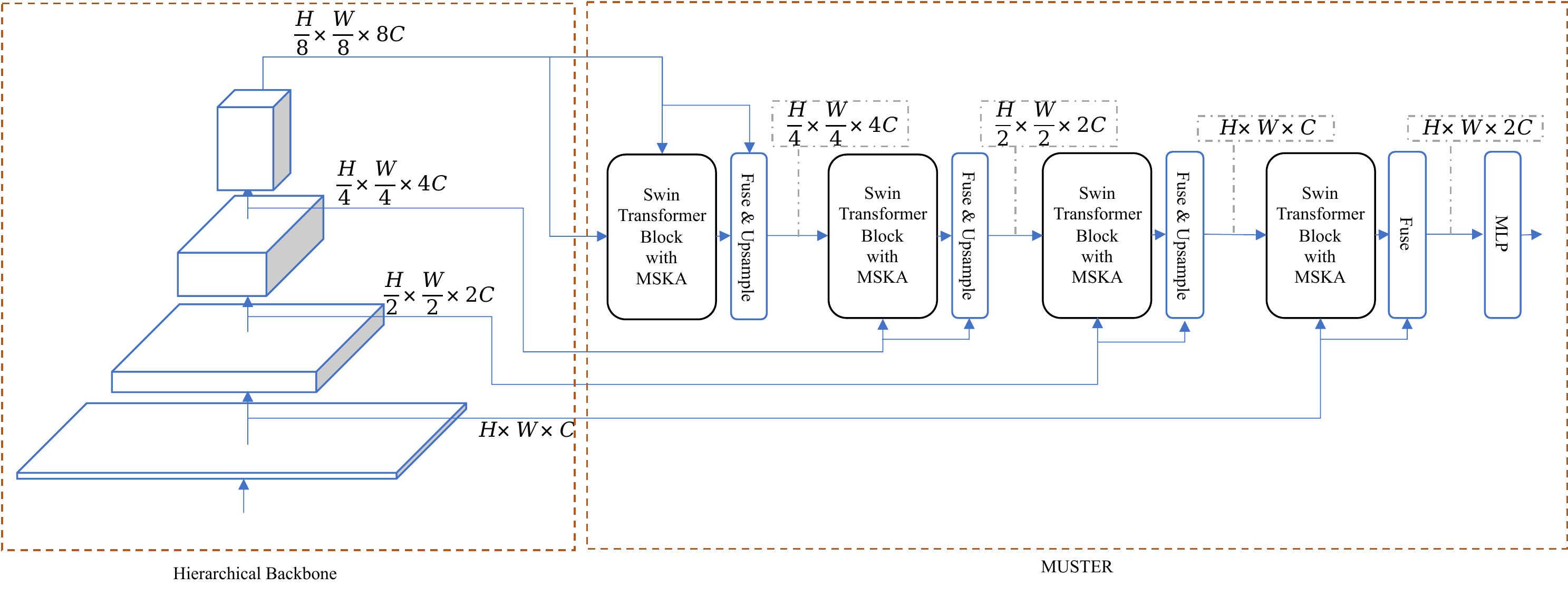}
    \caption{The left part is a pyramid backbone, which is not limited to the Swin Transformer. The right part is our proposed MUSTER, which is a multi-scale transformer-based decoder.  
    }
    \label{backbone}
\end{figure*}

\label{sec:formatting}
\subsection{Semantic Segmentation}
Semantic segmentation is an essential component in many visual understanding systems\cite{Survey,hao2022real}, which is a computer vision task that aims to assign each pixel in an image with a corresponding class label, which represents the meaning of the object or region it belongs to.   The goal of semantic segmentation is to enable machines to perceive and understand visual information in a way similar to human perception. 

Before the deep learning era, early attempts at semantic segmentation relied on low-level image processing techniques such as edge detection, thresholding, and clustering \cite{cheng2001color}.  
These methods often required hand-crafted features and were limited in their ability to capture high-level semantics.   One popular approach was based on graph-based optimization \cite{graph_optimization}, where an energy function was defined over the image pixels and optimized to obtain a globally consistent labeling. Another approach was based on Markov Random Fields (MRFs)\cite{MRFs}, where the label distribution was modeled using a probabilistic framework. However, these methods suffered from limitations in scalability, generalization, and accuracy, and could not handle complex scenes with multiple objects and occlusions. 
\subsection{Encoders of Semantic Segmentation}
Following the advent of AlexNet\cite{Alexnet}, both inchoate models like fully convolutional network\cite{FCN}, U-Net\cite{UNetCN}, SegNet\cite{SegNetAD} and recent strong models like Swin Transformer V2\cite{SwinV2}, SegFormer\cite{SegFormer}, SSformer\cite{shi2022ssformer} are all encoder-decoder based models. As a result, both encoder and decoder are of great significance in semantic segmentation. This work aims to propose a strong decoder that can make contribute to this field.

Since AlexNet\cite{Alexnet} was firstly proposed to apply convolutional network to get a outstanding performance in image classification, more and more CNN based models are designed. Fully convolutional network (FCN)\cite{FCN} is a classic method of semantic segmentation. This work is considered a milestone in image segmentation, since it demonstrated that deep networks can be trained for semantic segmentation in an end-to-end manner on variable sized images\cite{Survey}. There are many deep and effective convolutional neural architectures are proposed following closely, like HRNet\cite{HRnet} and ResNet\cite{resnet}. Spiking neural networks (SNNs)\cite{su2024multi,xu2021robust,xu2023constructing, shen2023esl, xu2018csnn}, as the brain-inspired neural networks, are also a strong architecture for semantic segmentation.

In recent years, as an effective method in natural language processing field, Transformer\cite{Attention} was adapted to the domain of computer vision, including image classification, semantic segmentation, etc\cite{PeterAFlach2015AdvancesIN}. Transformer is able to model the long-range dependency due to the self-attention mechanism. Vision Transformer (ViT)\cite{ViT} has convincingly proved that a pure transformer backbone can deal with all kinds of missions in computer vision. Since then, works based on Transformer has flourished, and some notable examples\cite{LVT,SegFormer,Swin,Twins,CSWinTA,ShuffleTR,LocalViTBL,TrainingDI,TokenstoTokenVT,mitunet,Longformer,EarlyCH} have arose. Among them, SETR\cite{setr} is a powerful work which firstly apply Vision Transformer on semantic segmentation and becomes state-of-the-art structure and Swin Transformer adds an inductive bias to make Transformer better adapt to pixel-level prediction tasks like semantic segmentation. These models have achieved state-of-the-art performance on various benchmark datasets and enabled a wide range of applications in fields such as autonomous driving, medical imaging, and robotics.
\subsection{Decoders of Semantic Segmentation}
Decoders for semantic segmentation are an essential component of encoder-decoder models.  One popular approach is the FCN-style decoder, which uses transposed convolutions to upsample the feature maps from the encoder and generate a segmentation map.   Another popular method is the PSPNet\cite{pspnet}, which incorporates a pyramid pooling module to capture multi-scale contextual information. DeepLab\cite{deeplabv3plus} employs atrous (dilated) convolutions to capture more extensive spatial context at multiple scales. 

Recently, transformer-based decoders have shown promising results in natural language processing tasks and have also been applied to computer vision problems such as object detection (e.g., DETR\cite{DETR}). However, the current status of transformer-based decoders for semantic segmentation is still in its early stages, with limited research and experiments conducted thus far. One challenge is the high computational cost associated with using transformers in dense prediction tasks like semantic segmentation.

\section{Method}
In this section we introduce MUSTER and Light-MUSTER, 
decoders designed for hierarchical backbones to better fuse the feature maps with different sizes 
and strengthen the attention mechanism in transformer. MUSTER retains detailed low-level information, 
and also gives good consideration to high-level semantic information. As a result, MUSTER achieves a great balance between segmentation details and semantic information.

\subsection{Overall Architecture}

An overview of MUSTER is demonstrated in Figure { \ref{backbone}. MUSTER requires the backbone to generate hierarchical feature maps as its inputs, and in our experiments, the C dimensions of these feature maps are geometric series. Actually, the channels are not strictly limited to geometric series since we can modify the channels while using $1\times 1$ convolution to upsample.

In each stage of MUSTER, we implement the output of the last stage and the feature map produced by the corresponding stage in backbone as the inputs. Both of them are used in the Swin Transformer Block with skip attention following skip-connection operation of Unet\cite{UNetCN}, which enables MUSTER to obtain more high-resolution information when upsampling. Then we pass both the output of Swin Transformer Block in decoder and encoder to upsample to keep the shape of output consistent to the next stage's feature map.

Supposing the four outputs of the backbone are $\{F_i(\frac{H}{2^{3-i}} \times \frac{W}{2^{3-i}} \times 2^{3-i} C) \mid i\in\{0, 1, 2, 3\}\}$, the process of every stage can be formulated as:
\begin{center}
\begin{equation}
\hat{F} = \begin{cases} SwinBlock(F_i, F_i) , i = 0 \\ SwinBlock(M_{i-1}, F_i) , i > 0 \end{cases} 
\label{Arichitecture Formula}
\end{equation}
\begin{equation}
M_i = \begin{cases} FuseUpsampleBlock(\hat{F}, F_i), i < 3 \\ FuseBlock(\hat{F},F_i), i = 3\end{cases}
\label{Arichitecture Formula 2}
\end{equation}
\end{center}

Each SwinBlock receives two feature maps for the proposed multi-head Skip Attention. We'll explain it in the next subsection.

Based on the proposed architecture presented in Figure \ref{backbone}, the computational complexity of MUSTER on an image of $h\times w$ patches is as follows (constant omitted):
\begin{equation}
\Omega(MUSTER)=hwC^2+M^2hwC
\label{complexity_eqn}
\end{equation}
where $M$ is fixed and is set to 12 by default. $M $ is the size of the window. The first term $hwC^2$ in the equation represents the computational complexity of MLPs in MUSTER. Since the computational complexity of self-attention for each window is $M^4C$, and there are $\frac{HW}{M^2}$ windows, the total self-attention complexity is $M^2hwC$.

The second term $M^2hwC$ represents the computational complexity of SW-MSKA in MUSTER. Since the computational complexity of window-wise attention for each window is $C$, and there are $M^2hw$ windows, the total window-wise attention complexity is $M^2hwC$.

As a result, the computational complexity of our proposed MUSTER is linear with the input image size, and it is still lightweight although transformer is applied here.
\begin{figure}[h]
    \centering
    \includegraphics[width=\columnwidth]{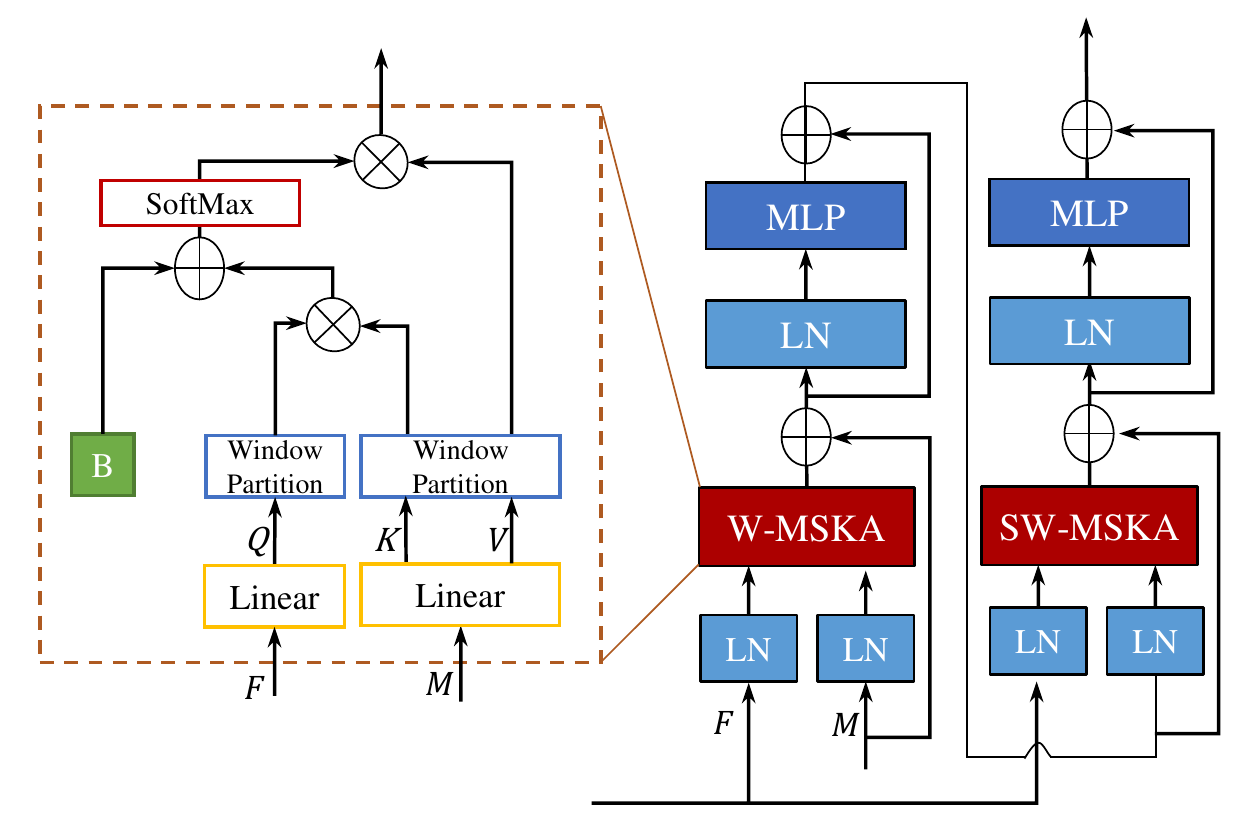}
    \caption{Multi-head Skip Attention Architecture. The shape of $Q, K, V$ should be $M^2 \times \frac{C}{N_{heads}}$, where $N_{head}$ means count head in Multi-head Skip Attention. }
    \label{fig3}
\end{figure}

\subsection{Multi-head Skip Attention}

In order to better utilize the feature maps from different stages of the backbone, we improve the attention mechanism used in Swin Transformer from multi-head Self Attention (MSA) to \textbf{M}ultihead \textbf{SK}ip \textbf{A}ttention (MSKA). We believe that it is difficult for the MSA to effectively activate a single category in one attention head since it simply calculates the similarity with itself. With MSKA, we calculate the similarity of two different feature maps with the same semantics, which better utilizes the attention mechanism and is able to activate a single category in one attention head.

Inspired by the interaction between the encoder and decoder of the Transformer used in the field of  natural language processing (NLP), we use the previous stage's output $M$ as Key-Value pairs, and the feature map $F$ as query. The Window-MSKA (W-MSKA) is estimated as:
\begin{center}
\begin{equation}
Attn(Q_F, K_M, V_M) = \mbox{SoftMax}(\frac{Q_F(K_M)^T}{\sqrt{d}} + B)V_M
\label{W-MSKA}
\end{equation}
\end{center}
where $Q_F$ is the query of Skip Attention, which is a linear transformation of $F$. $K_M$ and $V_M$ are key and value respectively, which are both linear transformations of $M$. $d$ is the dimension of $Q_F$ and $K_M$. $B$ is relative position bias mentioned in Swin Transformer\cite{Swin}.

Similarly, we regard the output of W-MSKA after layers of layer-norm and MLP as query, and the previous output $M$ as Key-Value pairs. The Shift-Window-MSKA (SW-MSKA) is estimated as:
\begin{center}
\begin{equation}
Attn(Q_{out}, K_M, V_M) = \mbox{SoftMax}(\frac{Q_{out}(K_M)^T}{\sqrt{d}} + B)V_M
\label{SW-MSKA}
\end{equation}
\end{center}

The complete process of the Swin Transformer Block with MSKA is presented in Figure \ref{fig3}.

\subsection{Fuse\&Upsample}
\label{fuse}
Contrary to other upsampling methods like interpolation or deconvolution, which rely on the limited information in a single feature map, we propose a succinct but effective way of upsampling by leveraging the complementary information contained in multiple feature maps from different sources.
The process of the Fuse\&Upsample block is presented in Figure \ref{fig4}.
\begin{figure}[htb]
    \centering
    \includegraphics[width=0.7\columnwidth]{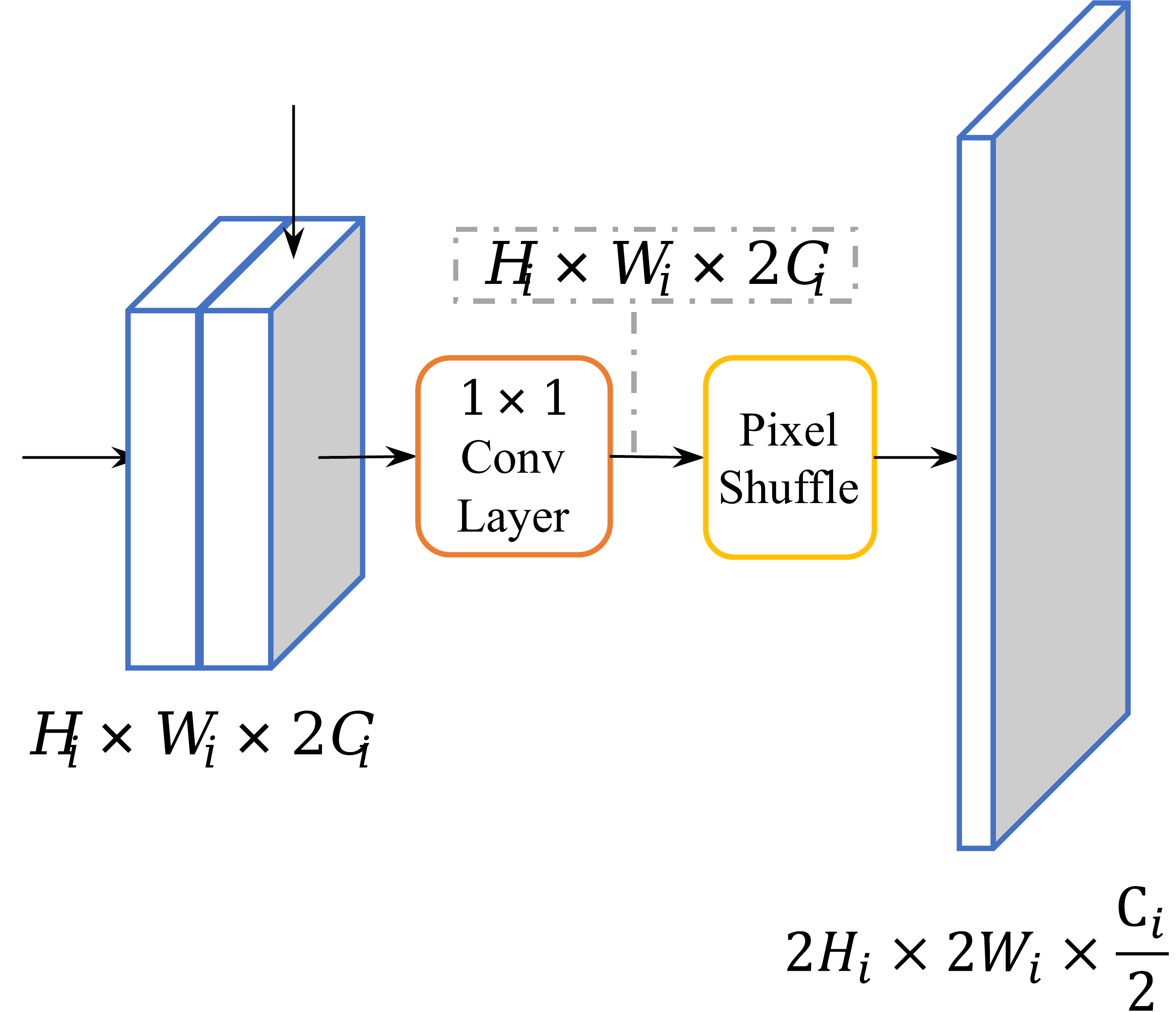}
    \caption{Fuse\&Upsample}
    \label{fig4}
\end{figure}

\begin{table*}[bth]
\begin{center}   
\caption{Detailed Training Configurations}
\label{table4} 
\begin{tabular}{llccc|lccc}
\toprule
Method & Backbone & Pretrain & Optimizer & \makecell{Learning\\Rate} & Dataset & \makecell{Batch\\Size} & Iterations & \makecell{Image\\Size}\\
\midrule
UperNet* & ResNet-101 & ImgNet1K\cite{JiaDeng2009ImageNetAL} & SGD\cite{optimizers} & 0.01† & \multirow{4}*{ADE20K} & \multirow{4}*{16} & \multirow{4}*{160000} & \multirow{4}*{$512\times512$}\\
MUSTER & ResNet-101 & ImgNet1K & AdamW\cite{adamw} & $6 \times 10^{-5}$ & & & \\
UperNet*& Swin-T-Base & ImgNet22K & AdamW & $6 \times 10^{-5}$† & & & \\
MUSTER & Swin-T-Base & ImgNet22K & AdamW & $6 \times 10^{-5}$ & &\\
\midrule
UperNet*& ResNet-101 & ImgNet1K & SGD & 0.01† & \multirow{3}*{Cityscapes} & \multirow{3}*{8} & \multirow{3}*{80000} & \multirow{3}*{$768\times768$}\\
MUSTER & ResNet-101 & ImgNet1K & AdamW & $6 \times 10^{-5}$ & & & \\
MUSTER & Swin-T-Base & ImgNet22K & AdamW & $6 \times 10^{-5}$ & &\\
\bottomrule
\end{tabular}   
\end{center}
*To make fair comparison with UperNet, we remove the FCN auxiliary head from these models and trained them by ourselves.
†The learning rates are in accordance with the configurations in their papers.
\end{table*}
We argue that an ideal upsampling operation -- by recovering high-frequency details -- should produce an output that contains more information than its low-resolution input does. Therefore, to perform upsampling effectively, additional cues is needed to supplement the information contained in the low-resolution input.

To achieve so, we first concatenate the output of Skip Attention Swin Transformer block with a feature map of the same scale in encoder, in order to enlarge channels from $C_i$ to $2C_i$. Then we use a $1 \times 1$ convolution to build a connection between the two combined feature maps. The $1 \times 1$ convolution can also be used to modify the channels to $2C_i$ if the feature maps from backbone and decoder have different numbers of channels. In the end, we apply Pixel Shuffle\cite{pixel_shuffle} to transfer the image scale from $H_i \times W_i \times 2C_i$ to $2H_i \times 2W_i \times \frac{C_i}{2}$. Pixel shuffle is estimated as:
\begin{equation}
\begin{aligned}
PS(T)_{x,y,c} = T_{\lfloor x/r \rfloor, \lfloor y/r \rfloor,c\cdot r\cdot mod(y,r)+c\cdot r\cdot mod(x,r)}
\end{aligned}
\end{equation}
In our model, $r$ equals to $2$. Eventually, the image scale is up-sampled from $\frac{H}{8} \times \frac{W}{8} \times 8C$ to the image scale $H \times W \times 2C$.
\begin{figure}[htb]
    \centering
    \includegraphics[width=0.7\columnwidth]{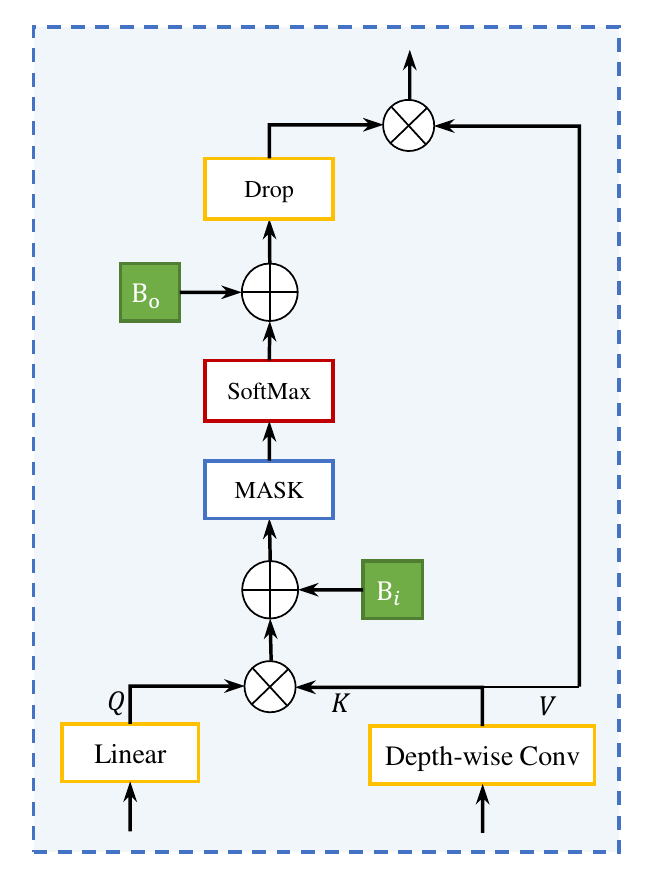}
    \caption{Forward process of attention mechanism in Light-MUSTER, where $B_i$ and $B_o$ represents learnable inner bias and outer bias.}
    \label{Light-MUSTER}
\end{figure}
\begin{figure}[htb]
    \centering
    \includegraphics[width=\columnwidth]{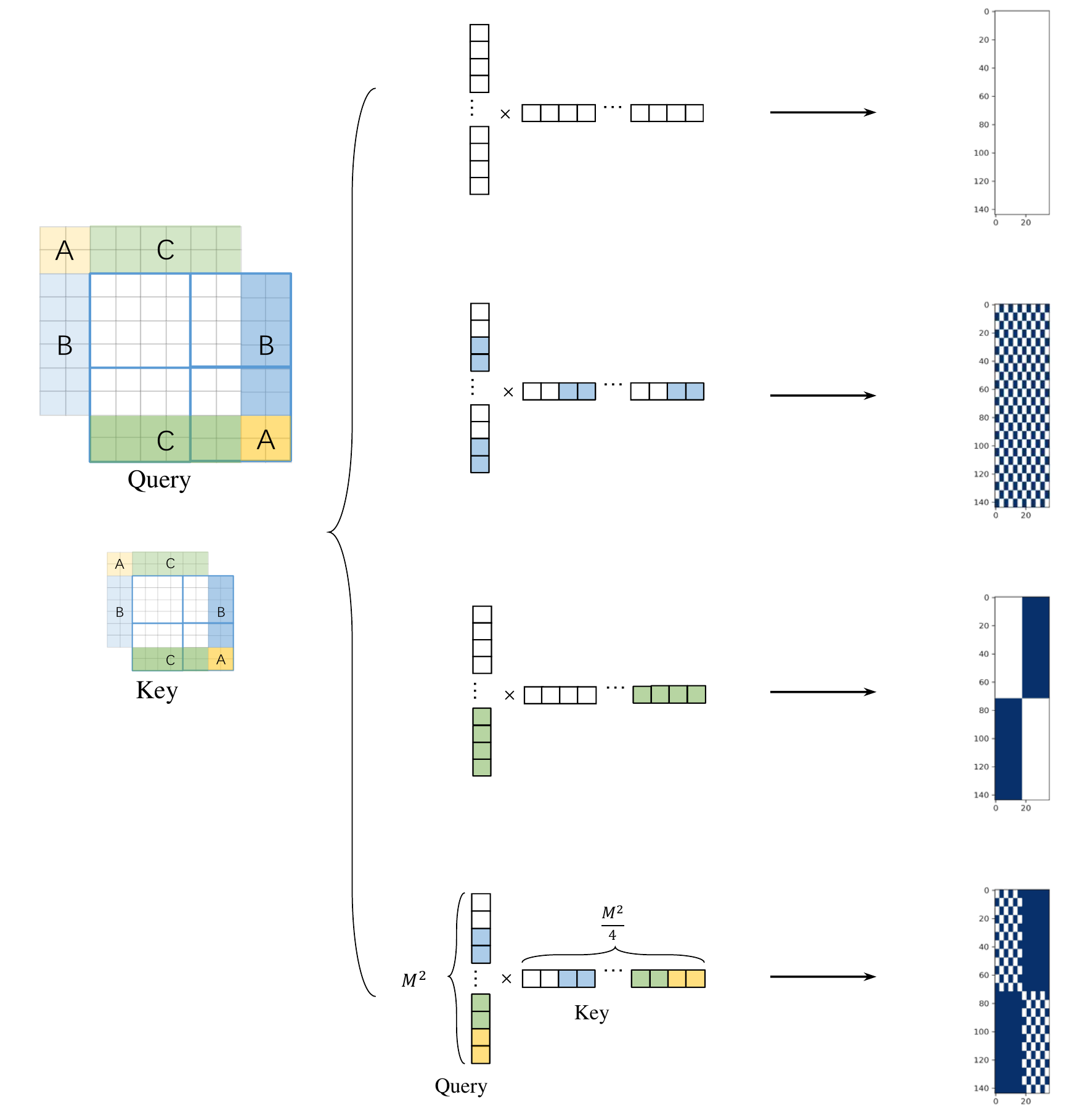}
    \caption{Transition of masks used for shifted window partition  in Light-MUSTER. The left image shows the method of shifting window applied on both query and key in scale of $M \times M$ and $\frac{M}{2} \times \frac{M}{2}$. The middle parts are the multiplication between flattened query and key with length of $M^2$ and $\frac{M^2}{4}$, respectively, and color of areas indicates the original patch where it belongs to. The right part presents the masks corresponding to the windows after cyclic shift. 
    }
    \label{transition}
\end{figure}
\subsection{Lightweight Attention Mechanism}
To make MUSTER lighter, we redesigned an attention mechanism for it which is used in Light-MUSTER, which is detailed described in Figure \ref{Light-MUSTER}. The whole process of our light attention mechanism can be formulated as:
\begin{equation}
\begin{aligned}
    Q &= \mbox{Linear}(X_{skip}) \\
    K, V &= \mbox{DepthwiseConv}(X)
\end{aligned}
\end{equation}
\begin{equation}
Attn(Q, K, V) = \mbox{SoftMax}(Q^TK + B_i + \mathcal{M} ) + B_o)V,
\end{equation}
where $\mathcal{M}$ means mask, which is applied in shift-window attention block.
There are three main modifications. Firstly, we removed superfluous linear layers in the forward process of the attention mechanism and added some learnable bias to maintain its learnability, i.e., the inner bias $B_i \in {\mathbb{R}}^{{M^2}\times{M^2}}$, and the outter bias $B_o \in {\mathbb{R}}^{{M^2}\times{M^2}}$, where $M$ represents the window size. Secondly, we used a depthwise convolution to downsample $K$ and $V$ in order to reduce FLOPS in the matrix multiplication. Thirdly, due to the change of shape of key and value, we had to redesign the original mask mechanism in Swin Transformer to fit it. The transition of four masks corresponding to the four areas in the shifted window is shown in Figure \ref{transition}, where
\begin{equation}
\mathcal{M}(x,y)=\left\{
\begin{array}{cc}
  0   & \text{ if $\mathcal{M}(x,y)$ is colored with white} \\
  -\infty   & \text{otherwise}
\end{array}\right.
\end{equation}

In the mask in Figure \ref{transition}, areas colored by dark blue mean the dot-product between the tokens is not necessarily computed due to from non-adjacent subwindows, so we need to mask the attention weights. The white parts indicate we calculate the self-attention  as in the non-shifted window. 

In implementation, it is complex to locate the original location of each window, so we used a hash map to recognize which new mask a window should match with.

\section{Experiments}

To verify the effectiveness of MUSTER and Light-MUSTER, we conducted a series of experiments and comparisons with other mainstream methods in the field of Semantic Segmentation. The results demonstrate the superior performance of our models.

Firstly, we compared our model with existing state-of-the-art methods for Semantic Segmentation. We evaluated the performance of MUSTER and Light-MUSTER on standard benchmark datasets and achieved remarkable FLOPS and results. These results highlight the effectiveness of our models in accurately segmenting semantic regions in images.

Furthermore, we extended the application of Light-MUSTER to weakly supervised semantic segmentation. This scenario poses additional challenges due to the limited availability of labeled data. However, Light-MUSTER successfully addressed these challenges and achieved impressive results even with weak supervision. This demonstrates the robustness and adaptability of our approach in handling different segmentation scenarios.

In addition to the comparisons and applications mentioned above, we conducted ablation experiments specifically targeting the MSKA and Fuse\& Upsample module. These ablation experiments provided valuable evidence of the importance of MSKA and Fuse\& Upsample module in achieving state-of-the-art performance in MUSTER.

\subsection{Datasets}
We conduct experiments using ADE20K~\cite{ade20k} and Cityscapes~\cite{cityscapes} datasets. ADE20K is a widely used semantic segmentation dataset, covering a broad range of 150 semantic categories. Compared to other datasets, scene parsing for ADE20K is challenging due to the huge semantic concepts. It has 25K images in total, with 20K for training, 2K for validation, and another 3K for testing. Cityscapes is a large-scale dataset which mainly contains images of urban roads, streets and landscapes. It has 5K images in total, with 2975 for training, 500 for validation, and another 1525 for testing.

\begin{table}[hbt]
\begin{center}   
\caption{MUSTER* Decoder Configurations}
\label{table6} 
\begin{tabular}{ccccc}
\toprule
Stage & \makecell{Output\\Downsp. Rate} &\makecell{Output\\Channels} & \makecell{Window\\Size} & \makecell{Attention\\Heads} \\
\midrule
Stage 1 & $16 \times$ & 512 & $12 \times 12$ & 32 \\
Stage 2 & $8 \times$ & 256 & $12 \times 12$ & 16 \\
Stage 3 & $4 \times$ & 128 & $12 \times 12$ & 8 \\
Stage 4 & $4 \times$ & 256 & $12 \times 12$ & 4 \\
\bottomrule
\end{tabular}   
\end{center}
*With Swin Transformer Backbone. The output channels are doubled when using ResNet-101.
\end{table}

\begin{table*}[htb]
\begin{center}   
\caption{Comparison of segmentation results of different methods on ADE20K and Cityscapes dataset. }
\label{table1} 
\begin{tabular}{llc|ccc|ccc}
\toprule
\multirow{3.5}*{Method} & \multirow{3.5}*{Backbone} & \multirow{3.5}*{Params} & \multicolumn{3}{c}{ADE20K} & \multicolumn{3}{c}{Cityscapes} \\
\cmidrule(lr){4-6}\cmidrule(lr){7-9}
& & & FLOPs** & mIoU & \makecell[c]{mIoU\\(ms+flip)*} & FLOPs & mIoU & \makecell[c]{mIoU\\(ms+flip)} \\
\midrule
FPN\cite{FPN} & ResNet-101\cite{resnet} & 47.51M & 65.04G & 39.35 & 40.72 & 520.33G & 75.80 & 77.40\\
UperNet\cite{Upernet} & ResNet-101 & 83.11M & 257.37G & 40.66 & 40.44 & 2361.16G & 78.84 & 79.9\\
\textbf{MUSTER}(Ours)† & ResNet-101 & 247.86M & 248.3G & 43.18 & 44.08 & 1895.92G & 79.94 & -\\
\midrule
FCN\cite{FCN} & ResNet-101-D8 &  68.59M	& 275.69G & 39.91 & 41.40 & 2205.48G & 75.52 & 76.61\\
DeepLabV3+\cite{deeplabv3plus} & ResNet-101-D8 & 62.68M & 255.14G & 45.47 & 46.35 & 2041.15G & 80.65 & 81.47\\
DMnet\cite{DMnet} & ResNet-101-D8 & 72.27M & 273.64G & 45.42 & 46.76 & 2188.49G & 79.19 & 80.65\\
PSPNet\cite{pspnet} & ResNet-101-D8 & 68.07M & 256.44G & 44.39 & 45.35 & 2051.15G & 79.77 & 81.06\\
PSANet\cite{psanet} & ResNet-101-D8 & 73.06M & 272.48G & 43.74 & 45.38 & 2178.44G & 79.69 & 80.89\\
SETR-PUP\cite{setr} & ViT-Large & 317.29M & 270.55G & 48.58 & 50.09 & Huge & 79.34& 82.15\\
SegFormer\cite{SegFormer} & MiT-B5 & 82.01M & 183.30G & 49.13 & 50.22 & 1460.4G & \textbf{82.25} & 83.48 \\
UperNet\cite{Upernet} & Swin-T-Base\cite{Swin} & 121.42M & 299.81G & 50.21 & \textbf{52.02} & 2352.35 & 82.00 & 83.28\\
SSformer\cite{shi2022ssformer}  & Swin-T-Base & 87.5M & 91.01G & 47.71 & - & 355.8G & 79.84&-\\
\textbf{MUSTER}(Ours) & Swin-T-Base & 138.94M & 139.53G & 50.18 & 51.8 & 1056.54G & 81.44 & 82.79\\
\textbf{Light-MUSTER}(Ours) & Swin-T-Base & 122.24M & \textbf{116.09G} & \textbf{50.23} & 51.88 & \textbf{949.01G} & 81.82 & \textbf{83.51}\\
\bottomrule
\end{tabular}  
\end{center}   
*ms+flip means we use multi-scaled and flipped data when testing. On ADE20K, the $512 \times 512$ images are augmented to $512 \times 512$ and $1024 \times 1024$. On Cityscapes, the $2048 \times 1024$ images are augmented to $2048 \times 1024$ and $4096 \times 2048$.\\
**On ADE20K, the FLOPs are calculated with input images of $512 \times 512$. On Cityscapes, the FLOPs are calculated with input images of $2048 \times 1024$. It should be noted that the performance of SegFormer presented in this table is not exactly the same as its original paper, because its original test scale is $640 \times 640$, but here we use 512 × 512 to control variable to meet hardware resource requirement.\\
†The maximum output channels of ResNet are 2048, while that of Swin-T are 1024. So the Params and FLOPs of MUSTER+ResNet are quite larger than MUSTER+Swin-T, but the FLOPs are still lower than most of the models.
\end{table*}

\subsection{Implementation Details}
We use MMsegmentation\cite{mmseg2020} as our codebase. Four RTX 3090 GPUs were used to train models for comparison experiments, while we only use a single RTX 3090 GPU for ablation studies. All experiments of MUSTER adopted AdamW as optimizer\cite{Adam, optimizers}, and the learning rate was set to $6 \times 10^{-5}$ in the beginning with a poly learning rate strategy and 1500 iterations' warm-up. The size of training images is $512 \times 512$ when using ADE20K, $768 \times 768$ when using Cityscapes. The swin transformer backbones are pretrained with Imagenet-22K. We test the models with both the same input size with the training images and resized, fliped images. Specific training configurations are presented in Table \ref{table4}. Attention heads means the number of heads in both W-MSA and SW-MSA in Swin Transformer, and that in both W-MSKA and SW-MSKA in MUSTER. Detailed configurations are expounded in Table \ref{table6}. The output channels should strictly correspond with the outputs of the hierarchical backbone, and Table \ref{table6} shows the case when using Swin-T-Base\cite{Swin} as the backbone. The window size and numbers of attention heads are the same with Swin-T-Base.

\subsection{Comparisons with the State-of-the-Art}

\begin{figure}[htb]
    \centering
    \includegraphics[width=\columnwidth]{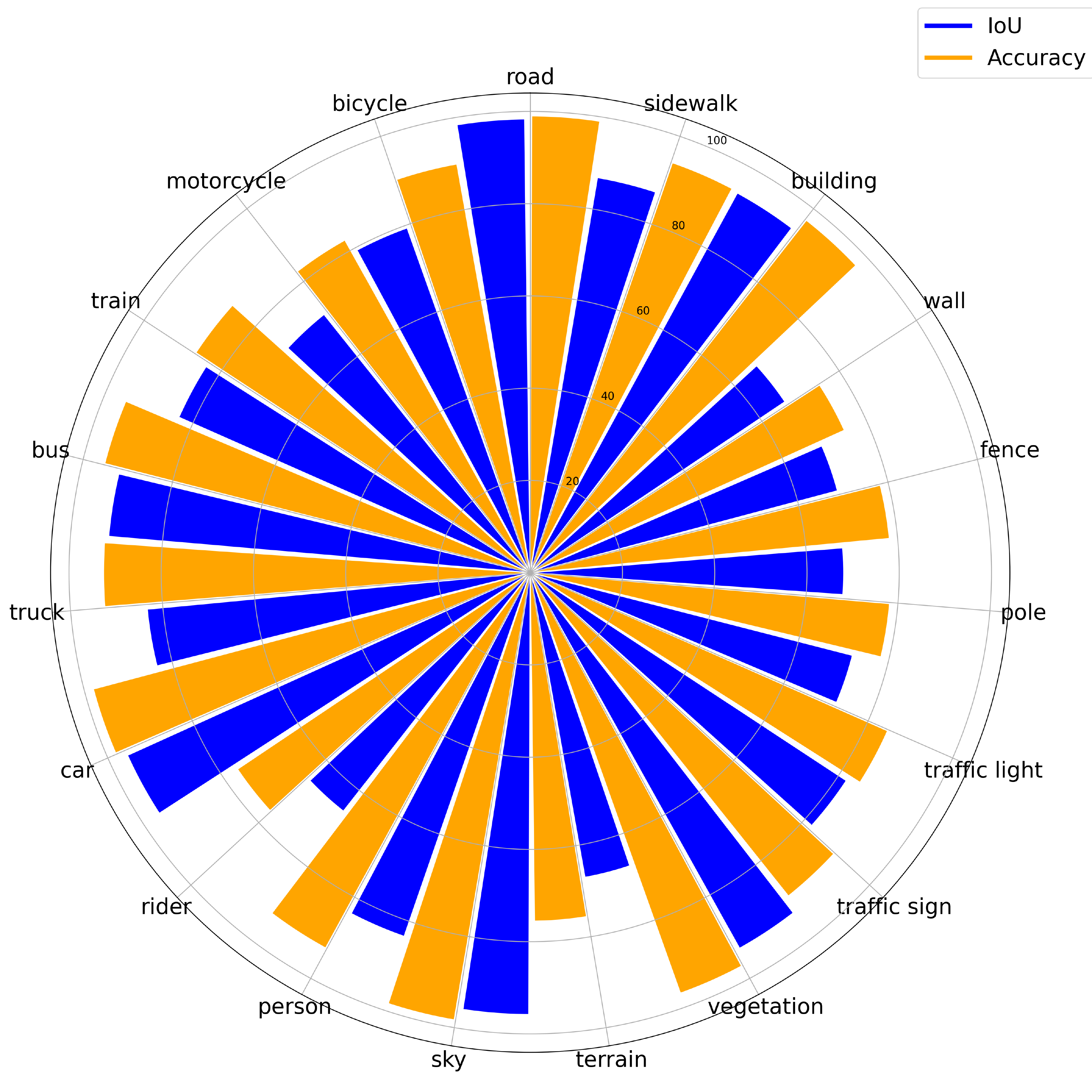}
    \caption{Swin-T-Base+MUSTER's mIoU and accuracy of each class in cityscapes.}
    \label{each_classs}
\end{figure}

\begin{table}[htbp]   
\begin{center}   
\caption{comparison of FPS on ADE20k.}
\label{FPS} 
\begin{tabular}{l|l|l|c}   
\toprule   Method & Backbone & FPS & mIoU \\  
\midrule
Light-MUSTER & Swin-T-Base & 11.2 & 50.23\\
MUSTER & Swin-T-Base & 11.0 & 50.18\\
UperNet & Swin-T-Base & 10.3& 50.21\\
SegFormer & MiT-B5 & 9.8 & 49.13 \\
SETR & ViT-Large & 5.4 & 48.58\\
DeeplabV3+ & ResNet-101 & 14.1 & 45.47\\
\bottomrule 
\end{tabular}   
\end{center}

\end{table}

\begin{figure}[htb]
    \centering
    \includegraphics[width=\columnwidth]{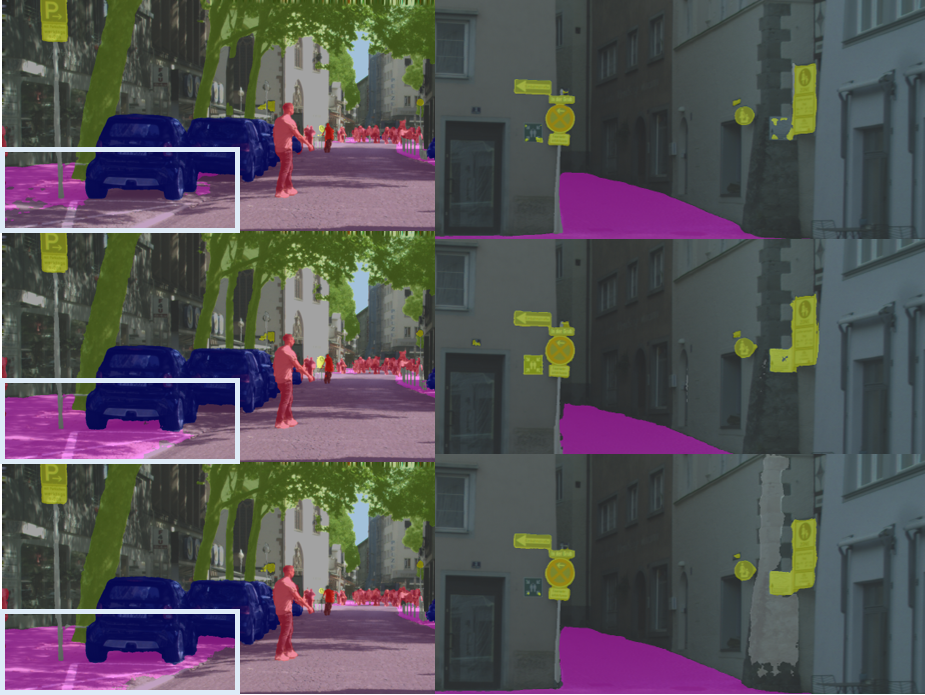}
    \caption{Comparison of segmentation results on Cityscapes. Above: UperNet+Swin-T-Base, middle: MUSTER+Swin-T-Base, below: Light-MUSTER+Swin-T-Base.}
    \label{City_comparison_fig}
\end{figure}

\begin{figure}[htb]
    \centering
    \includegraphics[width=\columnwidth]{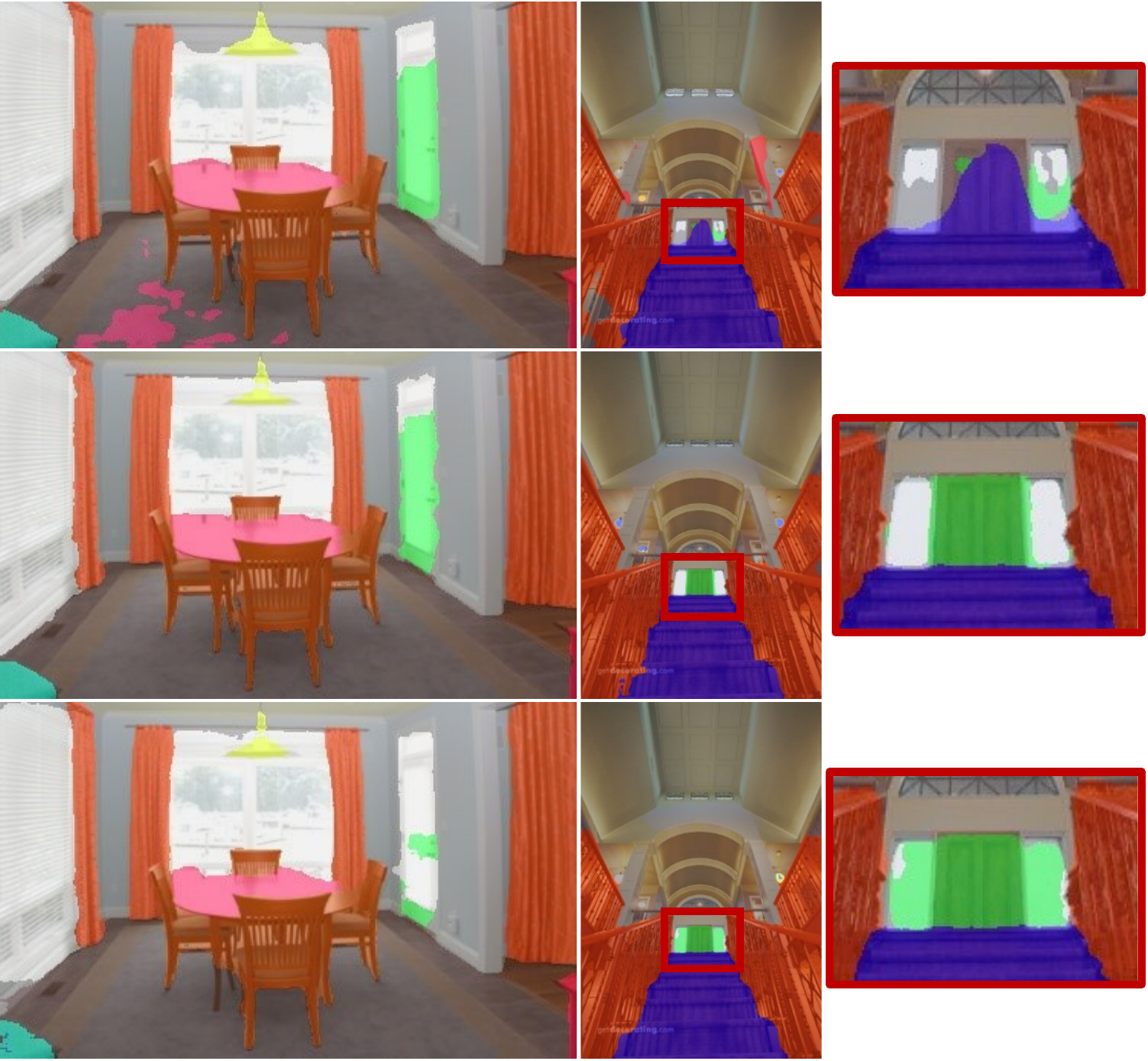}
    \caption{Comparison of segmentation results on ADE20K. Above: UperNet+Swin-T, middle: MUSTER+Swin-T, Below: Light-MUSTER+Swin-T.}
    \label{ADE_comparison_fig}
\end{figure}

\begin{figure*}[thb]
    \centering
    \includegraphics[width=\textwidth]{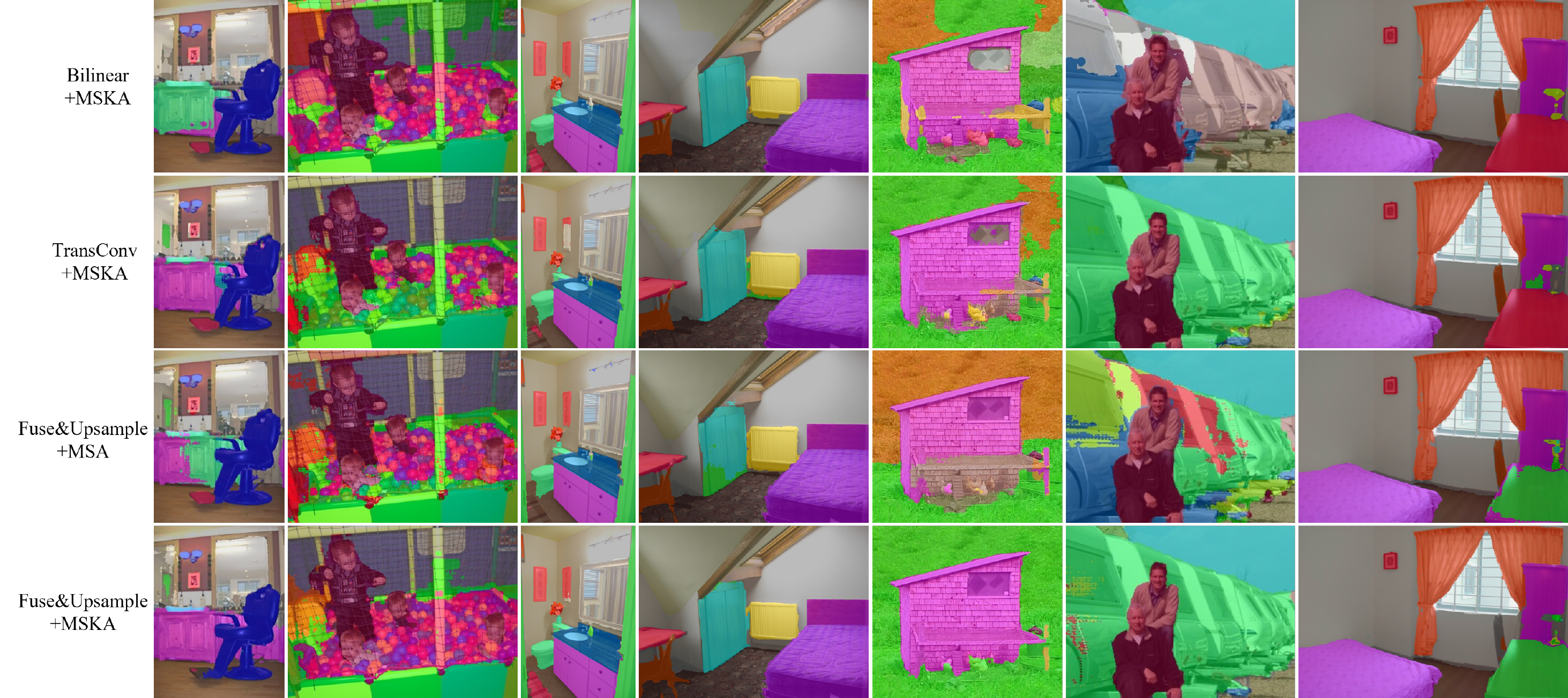}
    \caption{Visualization of segmentation results in ablation experiments.}
    \captionsetup{justification=centering} 
    \label{fig:Ablation Image}
\end{figure*}

As illustrated in Table \ref{table1}, on ADE20K, MUSTER with Swin Transformer backbone performs better than any CNN-based models from 10\% to 26\% while the FLOPs are 50\% lower. We believe that it is MUSTER's attention mechanism that makes MUSTER performs better when using CNN-based backbones. It illustrates that attention mechanism is effective not only in backbones, but also in decoders. Among transformer-based models, MUSTER achieves comparable performance with UperNet. Note that, UperNet is a multi-task learning framework, which is CNN-based and has multiple heads corresponding to various visual concept parsing, i.e., scene, object, part, material, texture. While, MUSTER only has one object segmentation head, whose FLOPs are less than half compared to UperNet. Moreover, MUSTER is distinguished by its relatively lightweight design, as its core components consist primarily of four Swin Blocks, in contrast to the more parameter-heavy MLP layers found in UperNet.Further, compared to recent specially-designed transformer-based models, SETR and SegFormer, MUSTER improves the mIoU metric by 1.65 and 1.1, respectively. SSformer\cite{shi2022ssformer} utilizes the same Swin Transformer backbone as MUSTER, paired with a lightweight all-MLP decoder. The superior results of MUSTER provide further insights into its performance advantages. The similar performance gains can be observed for multi-scaled and flipped data. 

As shown in Fig \ref{each_classs}, MUSTER's results of each class align well with human expectations of detection difficulty for various classes. There are no significant performance gaps, and the model performs strongly  in more challenging areas as expected.

We have also compared the FPS (Frames Per Second) with other mainstream models in \ref{FPS}. Our model shows an improvement of 8.7\% in FPS compared to the UperNet-based Swin Transformer.

On Cityscapes, since Cityscapes has less semantic
categories and is less challenging, the performance gap between old models and latest models is not as large as that on ADE20K, but the MUSTER’s advantage on efficiency of calculation still exists.

Figure \ref{City_comparison_fig} shows that MUSTER performs better when segmenting small objects like traffic signs. Besides, for large objects like trucks, The consistency of segmentation is also better than DeepLabV3+. Figure \ref{ADE_comparison_fig} also proves MUSTER's ability on segmenting small objects, but it sometimes causes jagged edges of segmentation under complicated semantic conditions, which decreases the final mIoU of MUSTER. We believe that the application of larger backbones or more MLP layers may fix this problem.

\begin{table}[htbp]   
\begin{center}   
\caption{Experiment results on PASCAL VOC 2012 val.}
\label{wsss} 
\begin{tabular}{l|l|c|c}   
\toprule   Method & Pseudo Label & mIoU \\  
\midrule
Light-MUSTER & RS+EPM & 75.71\\
RS+EPM & RS+EPM & 74.40\\
Light-MUSTER & RCA & 72.76 \\
RCA & RCA & 72.20\\
\bottomrule 
\end{tabular}   
\end{center}

\end{table}

\subsection{Application to Weakly Supervised Semantic Segmentation}
Furthermore, we tested our MUSTER on the task of weakly supervised semantic segmentation. Some segment results are shown in Figure \ref{wsss_fig}. In this task, we simply used pseudo labels produced by other models, e.g. RS+EPM\cite{RSEPM } and RCA\cite{ RCA }, to train our MUSTER without any additional modifications to the architecture. The results are shown in Table \ref{wsss}, which indicate that a pure Light-MUSTER can outperform some designs that include redundant loss or data interaction. This suggests that good-quality pseudo-labels are all that are needed for effective performance.
Combined with MUSTER, we are ranked the 2nd place at the time of submission in the leaderboard \cite{paperswithcode} of weakly supervised semantic segmentation on PASCAL VOC 2012 dataset. 
Figure \ref{wsss_fig} presents the weakly supervised segmentation results of employing Swin plus MUSTER as the encoder-decoder structure.

\begin{figure}[hbt]
    \centering
    \includegraphics[width=\columnwidth]{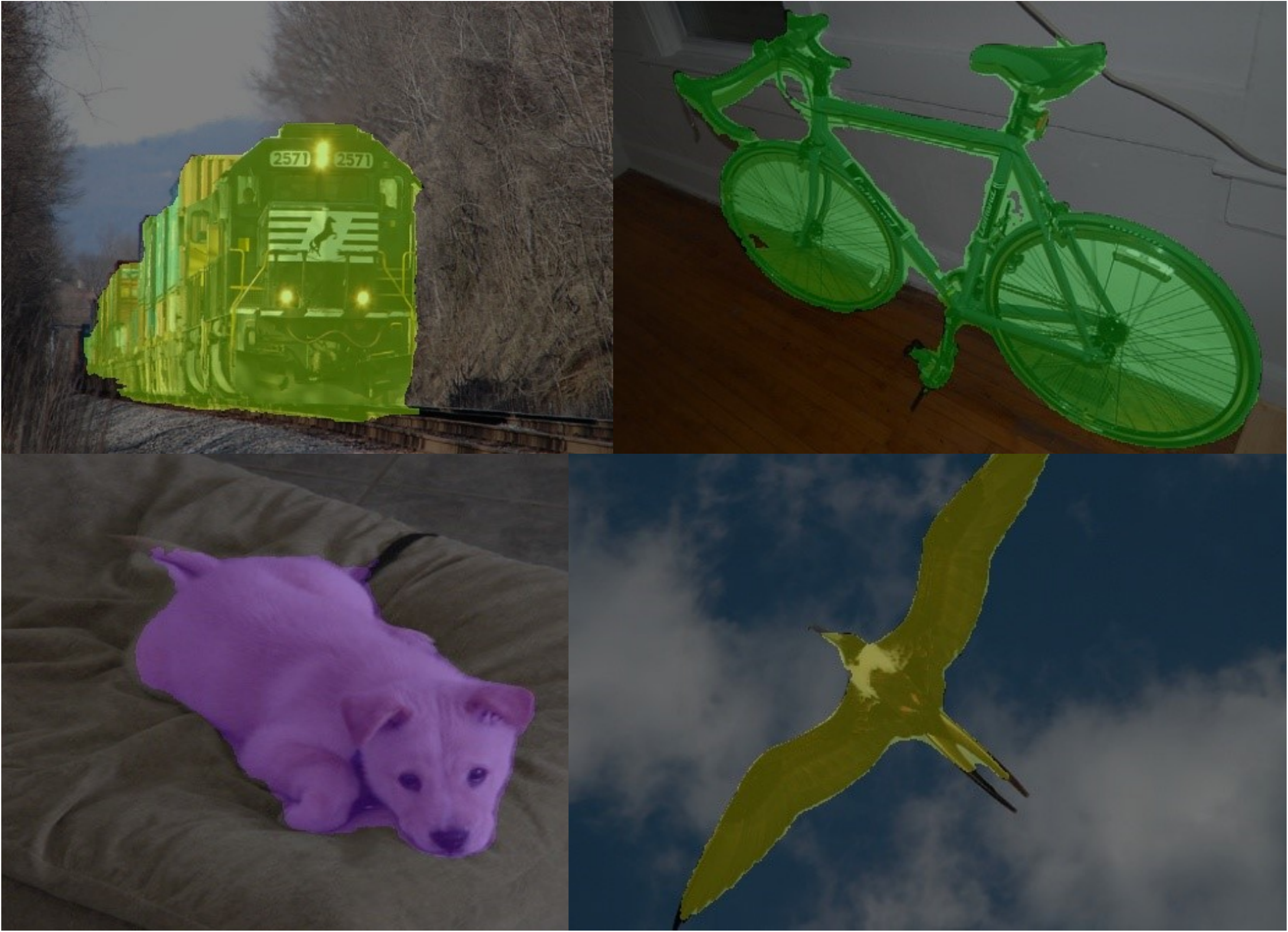}
    \caption{Visualization of segmentation results in PASCAL VOC 2012 validation set with weak supervision on the model architecture of Swin+MUSTER.}
    \label{wsss_fig}
\end{figure}

\subsection{Ablation Study}
\begin{table}[htbp]   
\begin{center}   
\caption{Ablation Experiment on MSKA}
\label{table2} 
\begin{tabular}{l|c|l|c|c}   
\toprule   Method & Backbone & Attention & Batch Size & mIoU \\  
\midrule
MUSTER & \multirow{3}*{Swin} & MSA & 4 & 46.54\\
MUSTER & & MSKA & 4 & 47.34\\
MUSTER & & MSKA & 16 & 50.18\\
\bottomrule 
\end{tabular}   
\end{center}
\end{table}

We also perform ablation studies to verify the effectiveness of the proposed MSKA. The results are presented in Table \ref{table2}. We make the ablation experiments in a smaller scale, which is still enough to demonstrate the effectiveness of MSKA. In Figure \ref{fig:Ablation Image}, we ablate the proposed key component visually, where we consider various combination between ``Bilinear, TransConv, and Fuse\&Upsample" and ``MSA and MSKA". By comparing ``Fuse\&Upsample+MSA" and ``Fuse\&Upsample+MSKA", we can observe the segmentation accuracy improvement  brought by MSKA block. Similarly, it is evident that Fuse\&Upsample can improve the results over traditional upsampling.    
\begin{figure}[hbt]
    \centering
    \includegraphics[width=\columnwidth]{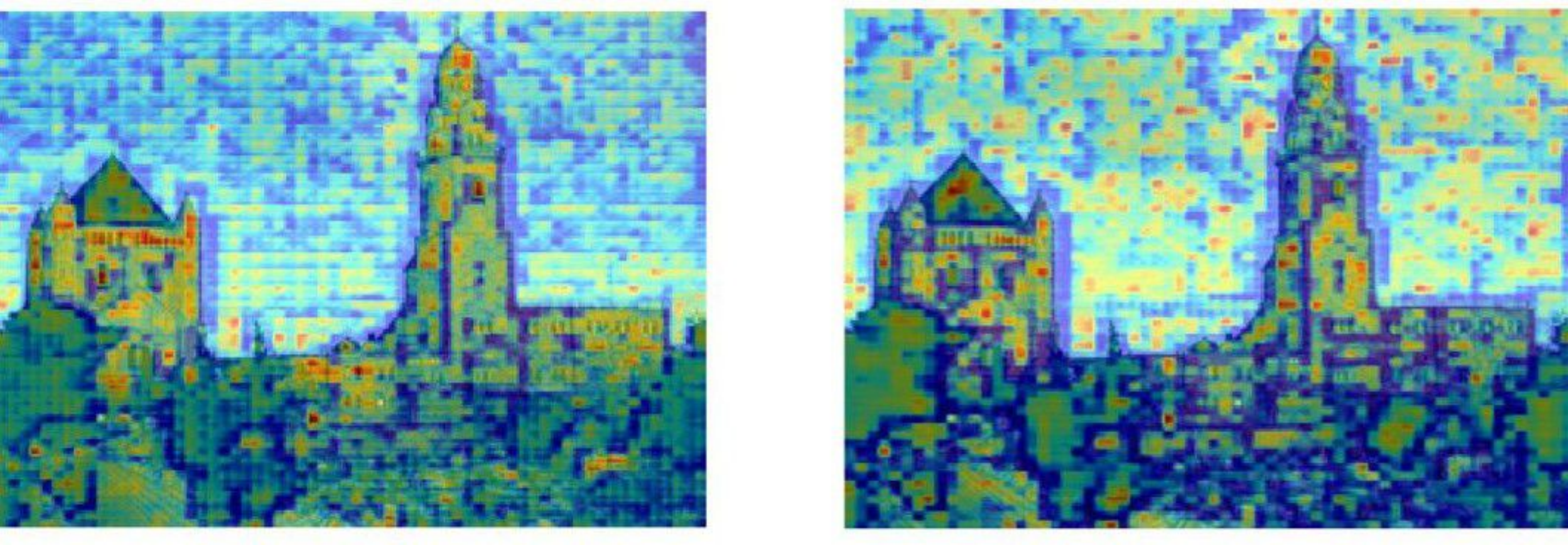}
    \caption{Comparison of feature maps between MSKA and MSA}
    \label{fig5}
\end{figure}
As shown in Figure \ref{fig5}, we compare the feature maps output from the third stage of decoder. Specifically, the images are fed to the same Swin Transformer backbone, but the left-hand one with proposed MUSTER with MSKA and the right-hand one with MUSTER with MSA instead. For both feature maps, we keep all architecture and training settings unchanged, except for the attention mechanism.
The image is chosen from validation data of ADE20K.
We can clearly infer from the feature map comparison that skip attention help to focus more on what we need, which are the tall buildings in the image.
\begin{figure}[h]
    \centering
    \includegraphics[width=\columnwidth]{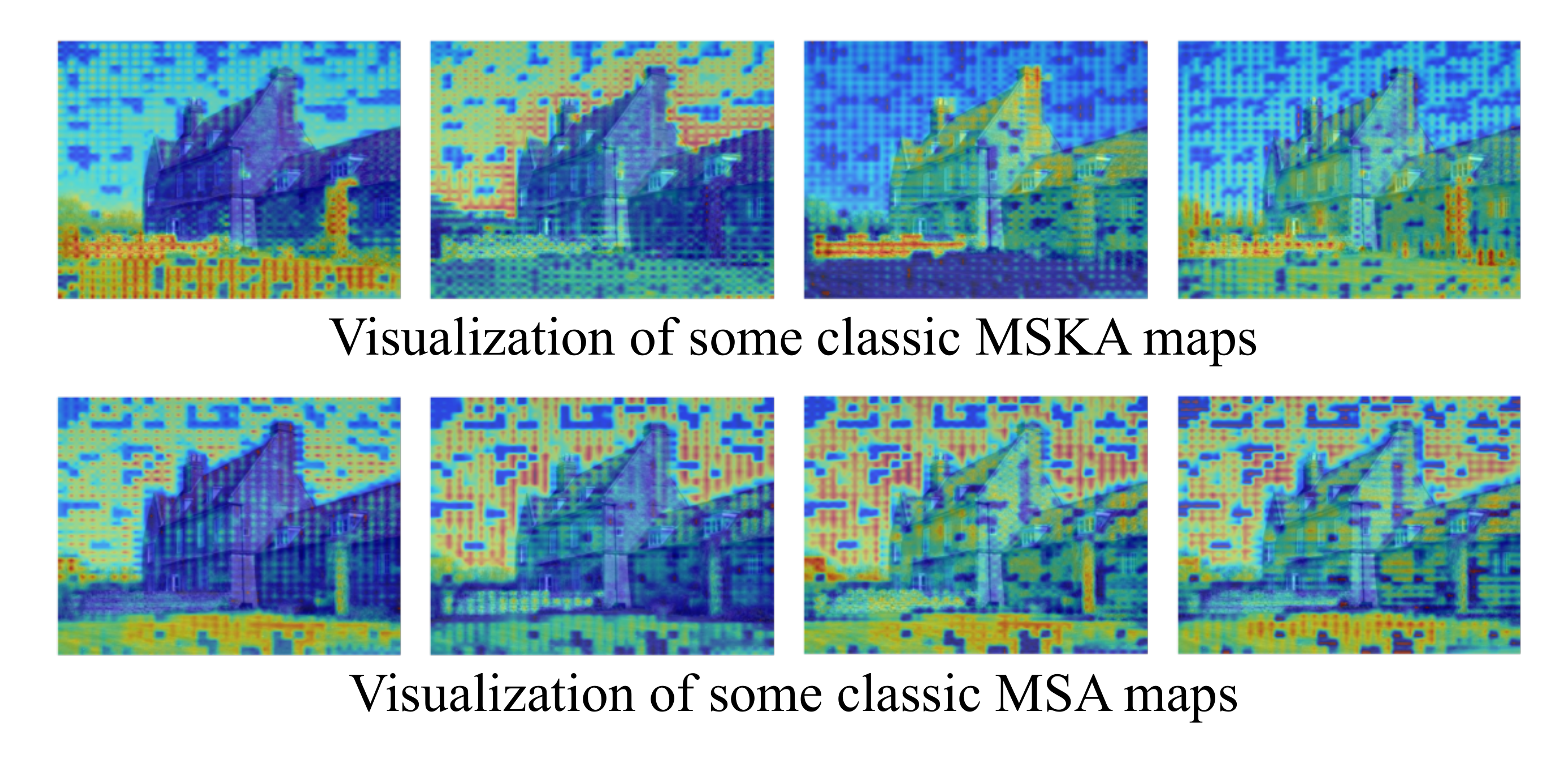}
    \caption{Visualization of feature maps extracted from models with MSA and with our MSKA.}
    \label{fig6}
\end{figure}
\begin{table}[htbp]   
\begin{center}   
\caption{Ablation Experiment on Fuse\&Upsample}
\label{table3} 
\begin{tabular}{l|l|c|c}   
\toprule   Method & Upsample Method & Batch Size & mIoU \\  
\midrule
MUSTER & Bilinear & 4 & 47.13\\
MUSTER & TransConv & 4 & 45.88\\
MUSTER & SelfConcat* & 4 & 46.22 \\
MUSTER & Fuse\&Upsample & 4 & 47.34\\
\bottomrule 
\end{tabular}   
\end{center}
*SelfConcat means that we concatenate two same feature maps that both from the last stage of the decoder, instead of what Fuse\&Upsample does. The other parts of SelfConcat are the same as Fuse\&Upsample.
\end{table}
Figure \ref{fig6} compares the output of different attention head from the second stage of decoder. It is obvious that in MSKA maps, one head of attention achieves better notice on a specific category such as the ground, the sky or the house. In different images, the same color represents the same value of  Meanwhile, the feature map of MSA is much weaker to differentiate the  objects of different categories.

Regarding ablation experiments of Fuse\&Upsample, we replace the concatenate operation with bilinear. It transforms the feature from $H_i \times W_i \times C_i$ to $2H_i \times 2W_i \times C_i$. Then we use a $1 \times 1$ convolution layer to transform the channels from $C_i$ to $\frac{C_i}{2}$.
In the experiments using TransConvolution, an layer of TransConvotion with kernal size $2 \times 2$ and stride 2 was applied to directly transform the feature from $H_i \times W_i \times C_i$ to $2H_i \times 2W_i \times \frac{C_i}{2}$ .
Table \ref{table3} illustrates that the proposed Fuse\&Upsample Block performs better than traditional upsample methods like bilinear and transconvolution. The experiment on SelfConcat proves that extra information from backbone does help improve model's performance.

\section{Conclusion}
In this work, we explore the way of designing a proper decoder that is able to effectively fuse features from the transformer encoder for semantic segmentation tasks. To achieve this, we present MUSTER, a fully transformer-based decoder, which comprises multi-head skip attention units. Meanwhile, we propose a enhanced while lightweight version of MUSTER, which is still powerful and is 18\% lighter than original MUSTER in FLOPS. We have done extensive experiments to demonstrate effectiveness and efficiency of the proposed approach on two publicly benchmarking datasets i.e., Cityscapes and ADE20K. In the future, we consider to apply MUSTER on some new architecture of semantic segmentation or some new segmentation tasks, e.g., maskformer and segment anything model, to achieve better performance.




%
\bibliographystyle{IEEEtran}
\bibliography{IEEEabrv,references}


 
\vspace{11pt}


\begin{IEEEbiography}[{\includegraphics[width=1in,height=1.25in,clip,keepaspectratio]{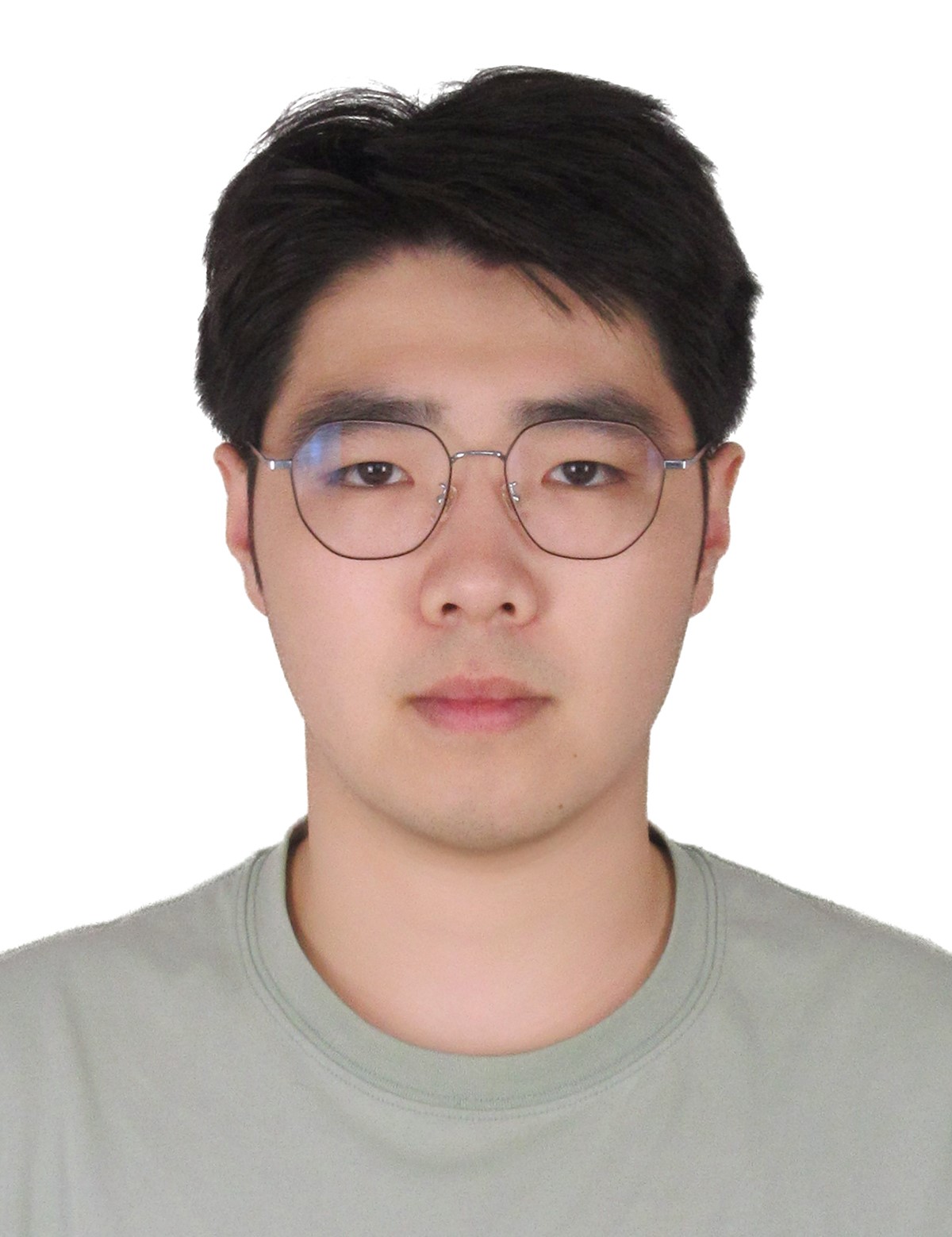}}]{Jing Xu}
is currently an undergraduate student in College of Computer Science and Technology in Nanjing University of Aeronautics and Astronautics, Nanjing, China. His current research interests include computer vision, representation learning and 3D vision.
\end{IEEEbiography}

\vspace{11pt}

\begin{IEEEbiography}[{\includegraphics[width=1in,height=1.25in,clip,keepaspectratio]{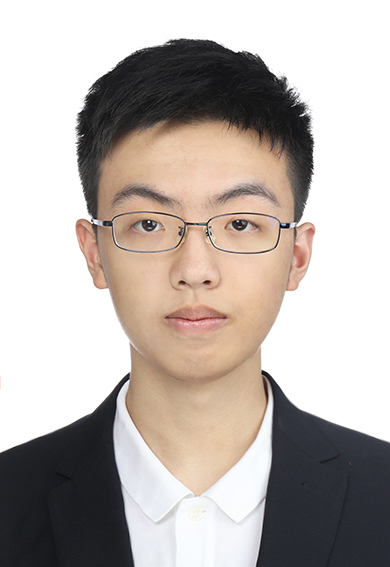}}]{Wentao Shi}
is currently an undergraduate student in school of computer science in Nanjing University of Aeronautics and Astronautics, China. His current research interests include computer vision and big data mining.
\end{IEEEbiography}

\begin{IEEEbiography}
[{\includegraphics[width=1in,height=1.25in,clip,keepaspectratio]{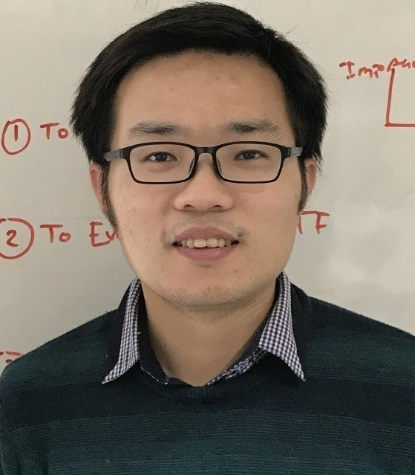}}]{Pan Gao} received  the Ph.D. degree in electronic engineering from University of Southern Queensland (USQ), Toowoomba, Australia, in 2017. Since 2016, he has been with the College of Computer Science and Technology, Nanjing University of Aeronautics and Astronautics, Nanjing, China, where he is currently an Associate Professor. From 2018 to 2019, he was a Postdoctoral Research Fellow at the School of Computer Science and Statistics, Trinity College Dublin, Dublin, Ireland.  His research interests include   computer vision, deep learning,  and artificial intelligence.
\end{IEEEbiography}

\begin{IEEEbiography}
[{\includegraphics[width=1in,height=1.25in,clip,keepaspectratio]{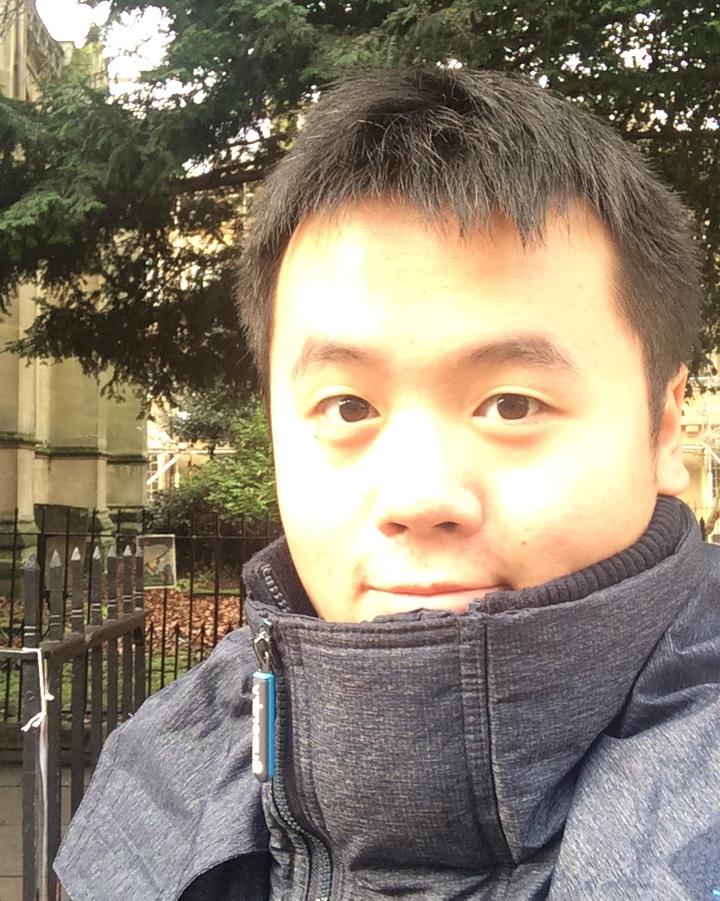}}]{Qizhu Li} is currently with TikTok Pte.~Ltd.~working primarily on recommendation algorithms. He completed his DPhil degree in Engineering Science with Philip Torr at the University of Oxford, where he focused on pixel-level object understanding at both category- and instance-level. Prior to that, he received his undergraduate degree from University of Oxford. His research interests include industrial-scale recommendation, object parsing and understanding, weakly and semi-supervised learning.
\end{IEEEbiography}

\begin{IEEEbiography}
[{\includegraphics[width=1in,height=1.25in,clip,keepaspectratio]{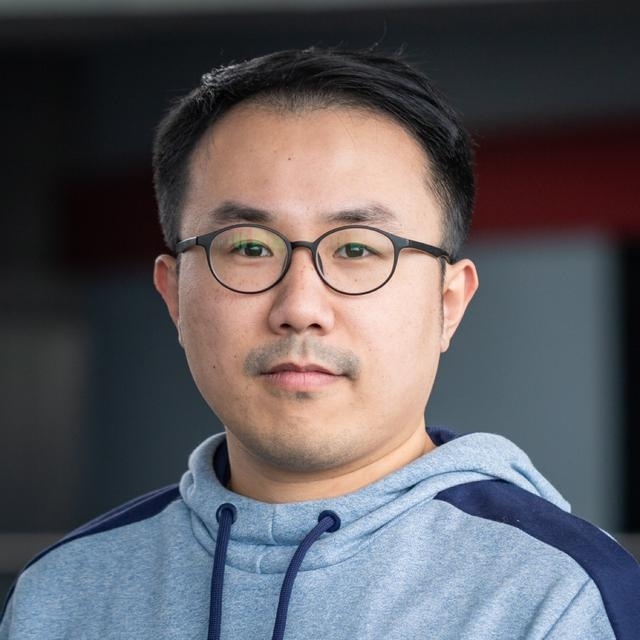}}]{Zhengwei Wang} is currently with ByteDance. Before that, he was a research fellow at the School of Computer Science and Statistics, Trinity College Dublin. He received his B.E. degrees in Electronic Engineering from Maynooth University 2014 and M.Sc. in Advanced Control System Engineering from the University of Manchester in 2015. He completed his Ph.D. at Insight Centre for Data Analytics, Dublin City University in 2019. Dr. Wang’ research interests span areas including video understanding, brain-computer interface, graph neural networks and large-scale recommendation systems. He also serves as a reviewer for several international journals such as Neural Networks, IEEE Transactions on Image Processing and IEEE Transactions on Emerging Topics in Computational Intelligence.
\end{IEEEbiography}

\vfill

\end{document}